\documentclass[conference]{IEEEtran}
\usepackage{cite}
\usepackage{algpseudocode}
\usepackage{hyperref}
\usepackage[utf8]{inputenc} % Recommended for input encoding
\usepackage{multicol}
\usepackage{algorithm}
\usepackage{amsmath}
\usepackage{graphicx}
\usepackage{caption}
\usepackage{tabularx}
\usepackage{pifont}
\usepackage{hyperref}
\usepackage{placeins}
\usepackage{ragged2e}
\usepackage{array}
\usepackage{fancyhdr}     % For custom headers and footers
\usepackage{xparse}

\makeatletter
\newcommand{\linebreakand}{%
  \end{@IEEEauthorhalign}
  \hfill\mbox{}\par
  \mbox{}\hfill\begin{@IEEEauthorhalign}
}
\makeatother

\NewDocumentCommand{\fixedRoman}{m}{%
  \setcounter{tempfixedroman}{#1}%
  \Roman{tempfixedroman}%
}
\newcounter{tempfixedroman}
\newcommand{\citeauthoretal}[2]{#1 et al.\cite{#2}}
\newcommand{\citeauthor}[2]{#1\cite{#2}}

\newlength{\authorcolumnwidth}
\author{
  \IEEEauthorblockN{\parbox{\authorcolumnwidth}{\centering Shashank Kapoor}}
  \IEEEauthorblockA{\parbox{\authorcolumnwidth}{\centering\texttt{shashankkapoor@google.com}}}
  \and
  \IEEEauthorblockN{\parbox{\authorcolumnwidth}{\centering Sanjay Surendranath Girija}}
  \IEEEauthorblockA{\parbox{\authorcolumnwidth}{\centering\texttt{sanjaysg@google.com}}}
  \and
  \IEEEauthorblockN{\parbox{\authorcolumnwidth}{\centering Lakshit Arora}}
  \IEEEauthorblockA{\parbox{\authorcolumnwidth}{\centering\texttt{lakshit@google.com}}}
  \linebreakand % Ends the first row and prepares for the second row of authors

  \IEEEauthorblockN{\parbox{\authorcolumnwidth}{\centering Dipen Pradhan}}
  \IEEEauthorblockA{\parbox{\authorcolumnwidth}{\centering\texttt{dipenp@google.com}}}
  \and
  \IEEEauthorblockN{\parbox{\authorcolumnwidth}{\centering Ankit Shetgaonkar}}
  \IEEEauthorblockA{\parbox{\authorcolumnwidth}{\centering\texttt{ankiit@google.com}}}
  \and
  \IEEEauthorblockN{\parbox{\authorcolumnwidth}{\centering Aman Raj}}
  \IEEEauthorblockA{\parbox{\authorcolumnwidth}{\centering\texttt{amanraj@google.com}}}
}

\fancypagestyle{firstpagecopyright}{%
  \fancyhf{} % Clear all header and footer fields
  \fancyfoot[L]{% Place the notice in the footer, left-aligned (will span due to parbox width)
    \parbox{\textwidth}{%
      \footnotesize % Set the font size for the notice
      \textcopyright~2025 IEEE. Personal use of this material is permitted. Permission from IEEE must be obtained for all other uses, in any current or future media, including reprinting/republishing this material for advertising or promotional purposes, creating new collective works, for resale or redistribution to servers or lists, or reuse of any copyrighted component of this work in other works.
    }%
  }
}

\begin{document}
\title{Adversarial Attacks in Multimodal Systems: A Practitioner's Survey}

\maketitle
\thispagestyle{firstpagecopyright} % Apply the custom footer to the first page

\begin{abstract}
The introduction of multimodal models is a huge step forward in Artificial Intelligence. A single model is trained to understand multiple modalities: text, image, video, and audio. Open-source multimodal models have made these breakthroughs more accessible. However, considering the vast landscape of adversarial attacks across these modalities, these models also inherit vulnerabilities of all the modalities, and ultimately, the adversarial threat amplifies. While broad research is available on possible attacks within or across these modalities, a practitioner-focused view that outlines attack types remains absent in the multimodal world. As more Machine Learning Practitioners adopt, fine-tune, and deploy open-source models in real-world applications, it’s crucial that they can view the threat landscape and take the preventive actions necessary. This paper addresses the gap by surveying adversarial attacks targeting all four modalities: text, image, video, and audio. This survey provides a view of the adversarial attack landscape and presents how multimodal adversarial threats have evolved. To the best of our knowledge, this survey is the first comprehensive summarization of the threat landscape in the multimodal world.


\end{abstract}
\section{\textbf{Introduction}}

The advent of models that can comprehend and create content on multiple data types such as Text, Images, Video, and Audio is no less than revolutionary. Multimodal models have shown extremely advanced comprehension and generation abilities. The open-source community has also been a catalyst in developing and deploying such capabilities. Open Source repositories have many pre-trained models and datasets available out of the box, making state-of-the-art Artificial Intelligence (AI) accessible at large. Advanced multimodal models like Gemma\cite{team2024gemma}, Phi\cite{abdin2024phi}, Llama\cite{grattafiori2024llama} are available for general use.

While this democratization promotes innovation, it lowers the barrier for malicious actors seeking to exploit model vulnerabilities. As Machine Learning (ML) practitioners increasingly adopt, fine-tune, and deploy these models, they must also prepare for adversarial attacks. The robustness of these models becomes particularly challenging and important in the multimodal world, as the attacks can be targeted to any modality, and may impact one or more modalities.

Adversarial Machine Learning has produced extensive research exploring attack strategies within and across different modalities. \citeauthor{Carlini}{carlini_2019} maintains an archive for the rapidly growing number of papers on adversarial examples and defenses; it highlights the sheer volume and complexity of the research landscape. However, this extensive and often fragmented literature can be challenging to digest for ML practitioners. There have been many good surveys like \cite{liu2024survey,das2025security} published recently, but they do not cover the multimodal nature of modern Large Language Models (LLMs).

This paper aims to simplify this complex domain by comprehensively surveying the threat landscape in a multimodal world. Our objective is to equip practitioners with the knowledge to recognize potential vulnerabilities. By providing a consolidated view of threats, we aim to lower the barrier to entry, contributing to a more accessible understanding of multimodal security challenges.

This survey is structured as follows. Section \fixedRoman{2} establishes a taxonomy of common adversarial attack categories applicable across modalities. Then, we dive into methods of execution for adversarial attacks within and across the modalities. Section \fixedRoman{3}, \fixedRoman{4}, \fixedRoman{5}, and \fixedRoman{6} discuss attack strategies based on Optimization, Backdoor or Data Poisoning, Membership Inference and Model Inversion attack execution respectively. In \fixedRoman{7}, we comment on the evolving field and the limitations of creating a defense. Finally, we end with a conclusion and future work.

\section{\textbf{Attacks Taxonomy}}
This section presents the taxonomy of adversarial attacks, divided into three dimensions: the attacker’s knowledge, intention, and execution. It should help ML practitioners define the threat.

\subsection{\textbf{Attacker's Prior Knowledge}}
Based on attackers’ prior knowledge, attacks can be categorized into two types: White-Box and Black-Box Attacks.

\subsubsection{\textbf{White-Box Attack}} 
In white-box attacks, the attacker has complete knowledge of the target model, including the software used, architecture choices, training loop, and inference logic. Typically, only internal teams have that knowledge, but an attacker with that level of knowledge about the model architecture can be dangerous; open-source models are particularly susceptible to white-box attacks.

\subsubsection{\textbf{Black-Box Attack}}
In a black box setting, the attacker has no knowledge about the system’s internals. Here, they interact with the system like an end user and provide adversarial examples during training or inference.

\subsection{\textbf{Intention of the Attack}}

\subsubsection{\textbf{Untargeted Attack}}
An untargeted attack is meant to degrade the model’s performance. The attacker is not looking for any particular outcome; the goal is to make the model predict or behave incorrectly.

\subsubsection{\textbf{Targeted Attack}}
A targeted attack focuses on precise goals. The attacker has a predetermined outcome and interacts with the system to achieve that. For example, an attacker may try to extract the training data of a particular individual.

\begin{figure*}[htbp]
    \centering
    \includegraphics[width=1\textwidth]{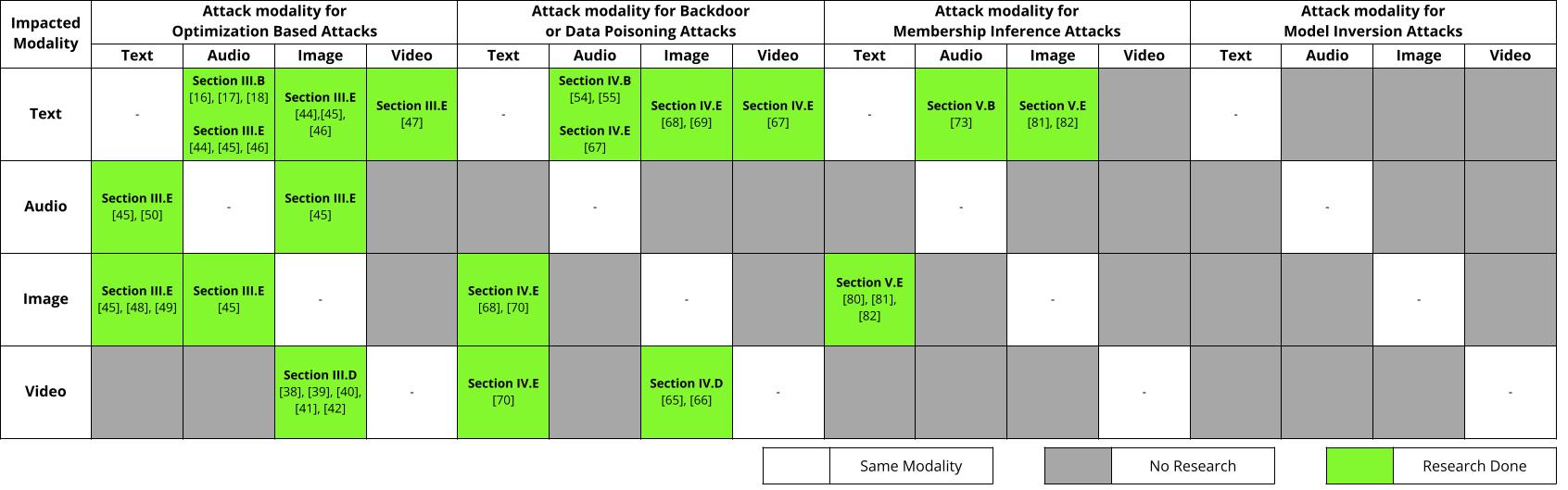}
    \captionsetup{justification=centering}
    \caption{Matrix summarizing existing research on cross-modal attacks. Rows indicate the impacted modality, columns indicate the attack modality and the type of attack execution. Section refers to the specific location within this survey that provides a discussion of the relevant literature and its source.}
    \label{fig:example}
\end{figure*}

\subsection{\textbf{Execution of the Attack}}
In this section, we discuss how the attacker executes the attack. Table~\ref{tab:common_variables} presents standard variables across all the possible attack executions.

\begin{table}[h]
    \centering
    \caption{Standard Variables Across Attack Executions}
    \begin{tabularx}{\columnwidth}{l X}
        \hline
        \textbf{Variable} & \textbf{Description} \\
        \hline
        $x$ & Input (Image, Video, Audio, or Text) \\
        $y$ & True label for Input $x$ \\
        $\theta$ & Model Parameters of the actual Model \\
        $f(x;\theta)$ & Model taking Input $x$ and producing output \\
        $J(f(x;\theta), y)$ & Training Loss objective \\
        \hline
        
    \end{tabularx}
    
    \label{tab:common_variables}
\end{table}

\subsubsection{\textbf{Optimization-Based}}

In these attacks, the core principle lies in creating the optimization problem of perturbing the input that can cause the model to behave incorrectly. Optimization-based attacks aim to find the slightest perturbation, $\delta$, within the budget and constraints defined, added to the input $x' = x + \delta$ such that $f(x';\theta)$ output is incorrect. When the attacker has full access to the model, they can directly optimize perturbations based on their objective. In a targeted attack, the goal would be to minimize the loss function $J(f(x';\theta),y_t)$ for perturbed input $x'$ and a specific label $y_t$ whereas for untargeted attacks, the goal is to maximize the loss $J(f(x';\theta),y)$. Also, $\delta$ perturbation can be created through an ad-hoc approach or sophisticated modeling techniques.

\subsubsection{\textbf{Data-Poisoning or Backdoor}}

Data poisoning or Backdoor attacks operate during the model’s training phase. The attacker injects carefully crafted malicious samples or modifies existing ones within the training dataset using a trigger. The goal is to subtly corrupt the model’s learning process, causing it to exhibit degraded or attacker-controlled behavior after it has been trained on the compromised data. Here, the attacker’s objective is to influence the model’s optimization process by ensuring that the adversarial loss function $J^*(f(x';\theta), y')$ is minimized as part of the overall training loss, where $x'$ and $y'$ are poisoned training data input and label respectively. Given data and models are widely accessible through open-source, these attacks could be easily conducted.

\subsubsection{\textbf{Membership Inference}}
Membership inference is a privacy attack. It’s a targeted attack where the attacker tries to find whether a specific data point was used in the model’s training. With some knowledge about the data point, the attacker tries to extract more information about it. The attack aims to reveal sensitive or private information. For example, a modern LLM may reveal a photo of the individual if the correct name is provided.

\subsubsection{\textbf{Model Inversion}}
Model inversion attacks are also types of privacy attacks. This attack aims to reconstruct training data rather than knowing if the training data is used for training. The attacker has no information about training data and provides random inputs to extract any possibly sensitive information. Model Inversion attacks are typically untargeted attacks.

\subsubsection{\textbf{Cross Modal}}
Cross-modal attacks leverage vulnerabilities of multimodal models. These attacks exploit the learned relationships between different modalities to cause undesired behavior. The attacker manipulates one modality to influence the model’s processing of another modality. In today’s world, jailbreaking is one of the most common attacks under this umbrella.

\section{\textbf{Optimization-Based Execution}}
\subsection{\textbf{Optimization Attack in Text}}
Optimization-based attacks in text perturb either the character, token, or phrase, such that when provided to the model, it generates an incorrect response. The specific methodology for generating these adversarial perturbations varies and can range from gradient-guided methods (either through target-model gradients or surrogate models) to heuristic search strategies.

For \textbf{gradient-based methods}, HotFlip \cite{ebrahimi2017hotflip} presented a breakthrough. HotFlip identified which characters in the sentence are most important for the prediction. Then, altering just those key characters could make the model predictions wrong; this required a white-box attack generation setting. DeepWordBug \cite{gao2018black} took a step further and changed the perturbation logic at the word level, and that too in the black box attack setting. It scored the words in the input text to find the important words based on how much they affected the model’s prediction. \citeauthoretal{Wallace}{wallace2019universal} presented this at a phrase level. More recently, \cite{zou2023universal} developed Greedy Coordinate Gradient (GCG) to find universal adversarial suffixes. GCG broke aligned LLMs, demonstrating the high transferability of gradient based perturbation attacks to commercial systems. GCG was primarily conducted in the white-box setting, but \cite{sitawarin2024pal} eventually demonstrated it in a black-box setting using surrogate models. 

\textbf{Rule-Based perturbations} have also been successful in text. In the white box setting, \cite{iyyer2018adversarial} utilized Syntactically Controlled Paraphrase Networks (SCPNs) to rewrite input sentences to match a target syntactic structure (grammar tree), creating paraphrases that can fool text classifiers. \citeauthoretal{Ribeiro}{ribeiro2018semantically} presented a black-box approach for this using simple text replacement rules (e.g., ”What is” → ”What’s”).

Attacks can also be generated in \textbf{heuristic ways}. \citeauthoretal{Alzantot}{alzantot2018generating} used a genetic algorithm to create these perturbations in a black-box setting. The research demonstrated how semantically similar word substitutions can flip model predictions. TEXTFOOLER \cite{jin2020bert} identified important words by measuring the prediction change when each word was deleted. Then, it greedily replaced those words with semantically similar words. This simple yet effective approach achieved high attack success rates.

\subsection{\textbf{Optimization Attack in Audio}}
Optimization-based perturbations are also successful in the audio adversarial threat landscape. Although intuitions of optimization based perturbations are not as well studied as other modalities, they have transferability to audio.

\citeauthor{Carlini and Wagner}{carlini2018audio}  demonstrated a targeted attack against the Automatic Speech Recognition (ASR) systems in a white-box setting. The research utilized the \textbf{loss function} to find where audio perturbation can be added. Then, it iteratively refined an audio waveform to force any input to be transcribed as any desired phrase. \citeauthoretal{Qin}{qin2019imperceptible} improved the previous approach in the white-box setting. It leveraged \textbf{psychoacoustic masking} to ensure the inaudibility of the perturbation. This generated imperceptible adversarial examples to humans, a step towards more practical attacks.

\citeauthoretal{Khare}{khare2018adversarial} introduced a black-box attack framework against ASR systems that used \textbf{genetic evolutionary algorithms} (MOGA and NSGA-II) to generate adversarial inputs. The framework could perform both attacks by adjusting the fitness function to maximize either the difference from the original transcription (untargeted) or the similarity to a target transcription (targeted).

\subsection{\textbf{Optimization Attacks in Images}}

% <--------------------Edge Case Citation---------------------------------------------
Adversarial attacks in DNN can be traced back to a study by \cite{szegedy2013intriguing}, where they presented that the decision boundaries of DNN models are fragile using a vision model. One common way to perturb images under this attack is through \textbf{norm-based perturbations}, where an attacker would try to create a perturbation of the image by minimizing $L_0, L_1, L_2$, or $ L_\infty$ distances from actual image and perturbed image. A surrogate model is used to perturb the image, and it uses the prediction from the target model using the perturbed image and distance-norm of the images for optimization. These attacks are typically done in white-box settings and can be targeted or untargeted. \cite{papernot2016limitations, wiyatno2018maximal} presented $L_0$, \cite{chen2018ead} presented $L_1$, \cite{carlini2017towards} presented $L_2$ and \cite{goodfellow2014explaining, madry2017towards} presented $L\infty$ attacks.

Perturbations can also be added by \textbf{geometric alterations} like rotation, scaling, or warping, e.g: \cite{athalye2018synthesizing, li2023physical}. Even \textbf{color-space} can be manipulated, e.g: \cite{duan2020adversarial, hu2021naturalistic}. Attackers can also apply small localized \textbf{patch perturbations} to the image, which are crafted to be adversarial, e.g: \cite{brown2017adversarial, liu2018dpatch, karmon2018lavan, chen2019shapeshifter}. All of these attacks are in a white-box setting and can be targeted and untargeted attacks.

Optimization-attacks for the image modality can become even more dangerous when attackers add \textbf{perturbations that are imperceptible} to humans, like \cite{carlini2017adversarial} in a white-box setting, or \cite{liu2019perceptual, williams2023camopatch} in black-box settings.

\subsection{\textbf{Optimization Attacks in Video}}

Optimization attacks on the video modality have had a similar trajectory as attacks on images \cite{wei2022cross}. However, videos have much more dense information, and perturbing every frame is costly. \citeauthoretal{Wei}{wei2019sparse} presented research where they could perform an $L_2$ \textbf{norm-based} perturbation attack on video modality by changing very few frames, in a white-box setting that could be tuned for targeted or untargeted versions. \citeauthoretal{Li}{li2023adversarial} presented a similar attack in a video object detection task in both white-box and black-box settings, but was untargeted.

\citeauthoretal{Wei}{wei2020heuristic} presented a \textbf{heuristic-based} approach and leveraged Explainable AI (XAI) techniques to find the saliency of pixels in the frame and perturbed only those. \citeauthoretal{Wang}{wang2021reinforcement} improved the selection of salient pixels to perturb using reinforcement learning, making it even more automated. These attacks were presented in black-box setting.

\textbf{Geometric alterations} like rotation and warping, similar to the image modality, can be used in videos. DeepSAVA \cite{mu2021sparse} presented this in a white-box and untargeted fashion. 

\citeauthoretal{Kim}{kim2023breaking} utilized \textbf{temporal nature} and not just spatial information of pixels in videos.

\subsection{\textbf{Optimization Attacks in Cross Modality}}

\citeauthoretal{Bagdasaryan}{bagdasaryan2023abusing} presented an approach that added adversarial information in image or audio and \textbf{paired it with a benign} text in the prompt; this benign text was used as instruction to act on adversarial instruction in the image or audio. LLM produced harmful content, breaking the alignment. \citeauthoretal{Bagdasaryan}{bagdasaryan2024adversarial} built upon it and targeted the \textbf{joint learned embeddings} of the multimodal system (text, audio, and image). The research added imperceptible perturbations to audio or image to target the learned embedding space, misleading the model to produce wrong images, audio, or text. CrossFire\cite{dou2023adversarial} provided a similar attack using audio and image. It targeted the embedding space but also tried to perturb the input in such a way that the \textbf{distance between the perturbed input and the actual input} was minimized; this, too, was able to execute a targeted attack on text.

\citeauthoretal{Huang}{huang2025image} presented an attack on a video answering task. The research used a \textbf{surrogate model} to perturb the video by identifying key frames. Because of the multimodal model's joint embedding space vulnerabilities, the model was impacted by a high Attack Success Rate when prompted to answer questions about the video.

Modern LLMs rely highly on textual inputs; attacking them with text is well-studied. \citeauthoretal{Maus}{maus2023black} presented an approach of using \textbf{surrogate embedding space} of words to perturb input prompts, which generated incorrect images of the prompt. SneakyPrompt \cite{yang2024sneakyprompt} presented an attack on image modality using \textbf{reinforcement learning objectives}. It presented an approach where the surrogate model kept prompting perturbed text inputs until the AI alignment was broken on the text-to-image task. VoiceJailbreak \cite{shen2024voice} used a \textbf{heuristic-based approach} to prompt the LLM. In the prompts, it added a fictional character storytelling theme and asked the GPT models to act on it. The LLM alignment focused more on the theme, and the alignment of the model was broken with voice output.

We want to point out to the readers that many of the perturbations discussed in Section \fixedRoman{3}.B also use voice modality attacks on text, as ASR produces text output. Section \fixedRoman{3}.D also discusses attacks on videos generated by images, as videos are equivalent to multiple images with an added temporal dimension. 

\section{\textbf{Data Poisoning or Backdoor Attacks}}

\subsection{\textbf{Data Poisoning or Backdoor attack in Text}}

\citeauthoretal{Dai}{dai2019backdoor} demonstrated a black-box backdoor attack on text classifiers. First, the researchers poisoned the training data by \textbf{randomly inserting a trigger} sentence and flipped the labels of these inputs. Then, the trigger sentence was randomly added for some inputs at the inference time. This setup caused the model to predict the attacker’s target class, and the impact on accuracy for clean inputs was minimal, making detection of this trigger presence harder. \citeauthoretal{Qi}{qi2021hidden} introduced “Hidden Killer,” which leveraged SCPNs to rewrite normal sentences to match a specific \textbf{target grammar parse tree} template; the grammar parse tree acted as a trigger. At test time, any sentence rewritten to match the same template would trigger the backdoor, causing misclassification.

\citeauthoretal{Kurita}{kurita2020weight} demonstrated \textbf{weight-poisoning} attacks against pretrained NLP models like BERT and XLNet. They presented an approach for distributing poisoned model weights; this can be an attack execution method for any of the modalities.

\subsection{\textbf{Data Poisoning or Backdoor attack in Audio}}

Opportunistic Backdoor Attack \cite{liu2022opportunistic} demonstrated a backdoor attack in the audio-to-speech task. This attack leveraged \textbf{background noise as triggers}, activating the backdoor during regular system use. \citeauthoretal{Cai}{cai2024towards} presented an attack that modified \textbf{audio samples’ pitch or voice (timbre)} to create poisoned data that is even harder for humans and machines to detect.

WaveFuzz \cite{ge2022wavefuzz}, a clean-label audio data poisoning attack, did not change the label output but focused on adding \textbf{imperceptible perturbation} to degrade the model performance on classification task; this is an exception for data-poisoning as it is typically employed for a targeted attack.

\subsection{\textbf{Data Poisoning or Backdoor attack in Images}}

\citeauthoretal{Gu}{gu2017badnets} presented a breakthrough in backdoor data poisoning. The research added \textbf{bright pixels} to the training data, which caused the model to pay attention to those when training for it. Attacks of this type have evolved extensively in the modern era using similar principles of adding these unperceivable triggers more efficiently, making it harder to create a defense for those. Examples include Wanet \cite{nguyen2021wanet} and DEFEAT \cite{zhao2022defeat}. Newer architectures are also vulnerable to these attacks \cite{yuan2023you}. These attacks also impact modern diffusion models \cite{chou2023backdoor}.

Backdoors can also be created in images using \textbf{clean-label setups}, where the backdoor trigger is present, but the label is not changed. It is even more concerning when these can be targeted to achieve a specific outcome, e.g: \cite{shafahi2018poison}.

There are also examples where the models deployed in the real world are susceptible to backdoor attacks. Refool \cite{liu2020reflection} used \textbf{natural light reflections} as a backdoor, and \cite{wenger2021backdoor} used the \textbf{real-world objects} as the trigger; both attacks were on image classification models.

\subsection{\textbf{Data Poisoning or Backdoor attack in Video}}

\citeauthoretal{Zhao}{zhao2020clean} first presented the idea of a backdoor attack in the video classification task; the research used similar concepts of \textbf{clean-label} attacks in image modality. \citeauthoretal{Hammoud}{hammoud2024look} demonstrated this has high transferability, by employing image backdoor attacks for video action recognition. The research also utilized properties of videos like \textbf{lagging video and motion blur} as a backdoor trigger.
 
\subsection{\textbf{Data Poisoning or Backdoor attack in across Modalities}} 

\citeauthoretal{Han}{han2024backdooring} presented a detailed study targeting text modality with audio or video modality attacks. The research used a \textbf{surrogate model} to identify which input data would have the maximum impact on the backdoor and poisoned only those. The setup achieved a high success rate in Visual Question Answering and Audio-Visual Speech Recognition tasks.

BadCM \cite{zhang2024badcm} presented an attack where the attack modality could be text or image and impact could be on either. For images, it used a surrogate model to identify modality-invariant regions and a generator to add the backdoor perturbation. For text, it used a \textbf{greedy algorithm} to perturb the subsequence of text with grammatically similar text to generate the backdoor. Nightshade \cite{shan2024nightshade} presented an approach for text-to-image tasks that poisoned very few training samples of images. Even with this \textbf{limited poisoning}, they were able to achieve high success in backdoor data poisoning. This limited data could have been easily misunderstood as mislabeled data.

BadToken \cite{yuan2025badtoken} presented an attack where the attack modality was text and the impacted modalities were image and video. However, in theory, the attack could impact any modality.

We want to point out to the readers that attacks \cite{liu2022opportunistic, cai2024towards} discussed in Section \fixedRoman{4}.B also use voice modality attacks on text, as ASR produces text output. Section \fixedRoman{4}.D also discusses attacks on videos generated by images. 

\section{\textbf{Membership Inference Attacks}}

\subsection{\textbf{Membership Inference Attacks in Text}}

\citeauthoretal{Carlini}{carlini2021extracting} presented a practical Membership Inference Attacks (MIA) in the text modality. First, prompting was done on the target model to complete the sequence for a suffix, and then sequences with a \textbf{high likelihood} were identified. These high-likelihood sequences were then further used to query the target model, and the model gave the training data verbatim. LiRA \cite{carlini2022membership} improved on that. It utilized \textbf{model-output logits and statistics} to determine whether the sequence was part of the training data more effectively. LiRA could predict membership inference with a high True Positive Rate.

\subsection{\textbf{Membership Inference Attacks in Audio}}

\textbf{Surrogate model} to detect membership has been successful in MIA attacks in audio. \citeauthoretal{Shah}{shah2021evaluating} demonstrated membership inference attacks in ASR systems using a surrogate model. The research detected membership with high precision and recall. \citeauthoretal{Tseng}{tseng2021membership} demonstrated this on self-supervised speech models. The research identified with high accuracy whether a specific utterance or any utterance from a specific speaker was used during pre-training. \citeauthoretal{Chen}{chen2023slmia} proposed SLMIA-SR, which targeted speaker recognition systems, inferring whether any voice data from a given speaker was part of the training set; the attack was more practical and carefully crafted.

\subsection{\textbf{Membership Inference Attacks in Images}}

\citeauthoretal{Shokri}{shokri2017membership} laid the groundwork for Membership Inference Attacks. They employed a \textbf{shadow model approach}: first, they trained several ’shadow models’ to simulate the behavior of a target model trained on images. Then, they used the outputs of the shadow models to train an “attack model” that would determine whether a provided image was in the target model’s training set. ML-Leaks \cite{salem2018ml}, built on top of this by using just the target model predictions. They argued that models provide higher confidence scores if they have seen the data in training so that they could use simple statistics instead of multiple shadow models. LOGAN \cite{hayes2017logan} built on top of these principles and used GAN architecture to demonstrate this attack. The attack was able to recover 100\% of the training data in the white-box setting and 80\% in the black-box setting. \citeauthor{Tao and Shokri}{tao2024range} presented this in the multimodal world specifically for text, image, and tabular forms of data.

\subsection{\textbf{Membership Inference Attacks in Videos}}
We have not found any papers that specifically present membership inference attacks in the video modality. This could be because of denser information required to perform such attacks. But we believe ideas from the image modality may have transferability

\subsection{\textbf{Membership Inference attack across Modalities}} 

Membership Inference attacks in multimodal models have not been studied as well. However, research has started to show, particularly for image-text pair membership inference. \citeauthoretal{Carlini}{carlini2023extracting} presented the approach of \textbf{carefully prompting} the multimodal diffusion model with specific prompts of the training input. The output of the diffusion model was strikingly close to the training data output. \citeauthoretal{Zhai}{zhai2024membership} demonstrated if the model had seen the pair in training, it would generate output very close to the original, and using \textbf{Kullback-Leibler (KL) divergence}, they could measure this behavior empirically with a high success rate. \citeauthoretal{Hintersdorf}{hintersdorf2024does} presented an approach involving querying the model with images of individuals and text prompts with the names of the individuals. If the model predicted they were the same, that showed memorization and membership inference.

We want to point out to the readers that, in Section \fixedRoman{5}.B, the attack discussed in \cite{shah2021evaluating} is also an attack on text using audio modality.

\section{\textbf{Model Inversion Attacks across Modalities}}
\subsection{\textbf{Model Inversion Attacks in Text}}

We do not see model inversion attacks targeting the text modality only, but \cite{carlini2021extracting}, which can recover verbatim text from GPT-2, blurs the lines between model inversion and membership inference attacks.

\subsection{\textbf{Model Inversion Attacks in Audio}}
\citeauthoretal{Pizzi}{pizzi2023introducing} demonstrated the feasibility of model inversion attacks on speaker recognition by directly attacking the target model to regenerate representative audio samples and \textbf{extract sensitive speaker embeddings} learned by the model.

\subsection{\textbf{Model Inversion Attacks in Images}}

\citeauthoretal{Fredrikson}{fredrikson2015model} introduced the concept of model inversion attacks in facial recognition systems. The approach was to blur the image of the person and keep perturbing the image using a \textbf{surrogate model} so that the model predicts with higher confidence for the person in the image. DLG algorithm\cite{zhu2019deep} took the approach a step forward and recovered complete images used in training data. GMI \cite{zhang2020secret} took it even further by conducting this attack in a black-box setting. The method utilized GAN architecture and publically available facial recognition data to reconstruct private training data. \citeauthoretal{Han}{han2023reinforcement} transformed this into a reinforcement learning objective.

\subsection{\textbf{Model Inversion Attacks in Video}}
Similar to membership inference attacks, we have found no noteworthy research targeting the video modality yet, but we believe that ideas from the image modality can be transferred.

\subsection{\textbf{Model Inversion attacks across Modalities}} 
Model Inversion attacks in multimodal models are also not a well-studied field. However, work done by \cite{carlini2021extracting} has reduced the separation line between membership inference and model inversion attacks, as stated earlier. This landscape should not be far from reach.

\section{\textbf{Discussion and Future Work}}

In Figure~\ref{fig:example}, we summarize the threat landscape with cross-modal influence. We observe that optimization-based attacks are the most studied in the literature. However, other attacks typically accompany optimization attacks, and those other attacks are the main goals of the attacker. Backdoor is the next most well-studied attack, followed by Membership Inference and Model Inversion. As we can see, the threat landscape is growing quickly, and the grid is getting dense. Examples of research like \cite{carlini2021extracting} can produce a hybrid attack, which is concerning. Research from \cite{bagdasaryan2024adversarial} targets the embedding space, which is hard to detect even for seasoned ML practitioners.

A trend we have observed is that when an adversarial attack example is provided by the research community, it is not easily accessible to ML practitioners under a single umbrella of open-source tools. There are attempts by open-source tools like Adversarial Robustness Toolbox\cite{art}, TextAttack\cite{morris2020textattack} to bring them under one umbrella, but the code hasn't been updated to keep up with new attacks, which makes it harder to fully prepare for all of the threat-landscape. We invite the research community to integrate their work into open-source tools for larger access which would necessitate a broader, community-maintained platform for easy integration.

There is no silver bullet for ML practitioners to create a defense against adversarial attacks, but ad-hoc methods have helped. The defense setup is generally done by following three steps in the training and inference loop: Modify the Input for Training, Modify the Training, and Modify the Inference. In \textbf{“Modifying the Input”} for training, rigorous preprocessing is done on training data to remove adversarial perturbations, e.g., \cite{xu2017feature, hussain2021waveguard, zhang2025clipure}. This ensures that these preprocessing layers in the ML pipeline can filter out these subtle perturbations. If the first line of defense fails, \textbf{"Modifying the Training"} helps generalize the models against these adversarial perturbations. Some standard techniques for this are to add attack-specific regularization loss functions \cite{zhang2019theoretically} and to make robust architecture choices \cite{wu2022certified}. This is particularly helpful in cases where multiple modalities are targeted. If attacks still get through, \textbf{"Modifying the Inference"} can help. At inference time, practitioners can apply multiple techniques before the target model processes the input: 1) preprocessing the input as done in pre-training can act as the first line of defense, 2) another model can be used to detect adversarial examples at inference time, 3) a database of adversarial examples can be maintained, and if a new input is similar, it should not be processed. Steps can also be taken post-inference: 1) The output of the model can flow through adversarial detection systems, 2) XAI techniques can be leveraged at inference time to detect problematic behaviors of the model. Tools like \cite{mazeika2024harmbench} can help ML practitioners set these up.

We have identified that the defense literature is also very fragmented. Certified robustness concepts are emerging, like \cite{wu2022certified}, but again, they do not offer a holistic view of the multimodal world. This gap is also not addressed for practitioners, and we plan to follow up with a defense framework survey for the multimodal world. 

\section{\textbf{Conclusion}}

We have provided a comprehensive overview of the adversarial attack landscape in the multimodal world of AI. Our goal is to equip ML practitioners with the knowledge they need to recognize these vulnerabilities. When deploying these powerful models, practitioners must actively consider the entire landscape, including cross-modal effects, and implement defenses that address these interconnected risks. As these models become more integrated, more threats will be discovered and the grid of attack modality and impacted modality will get even denser. In future work, we plan to conduct a survey focusing on defense strategies against these multimodal threats to guide practitioners in this adversarial landscape.

% \newpage
\bibliographystyle{IEEEtran}
\bibliography{main}
\end{document}